\documentclass[letterpaper, 10pt, conference]{ieeeconf}      

\IEEEoverridecommandlockouts                              

\usepackage{graphicx} 
\usepackage{epsfig} 
\usepackage{epstopdf}
\usepackage{tikz}
\usepackage{mathptmx} 
\usepackage{times} 
\usepackage{amsmath} 
\usepackage{amssymb}  

\usepackage{algorithm}
\usepackage{algorithmic}

\usepackage{tablefootnote}
\usepackage{color}
\usepackage{amsfonts}
\usepackage[noadjust]{cite}
\usepackage{subfigure}
\usepackage{threeparttable}
\usepackage[framed,numbered,autolinebreaks,useliterate]{mcode}
\usepackage{soul}
\usepackage{tasks}
\usepackage[normalem]{ulem}
\usepackage{pgf}
\usepackage{tikz}
\usetikzlibrary{arrows,automata}

\usepackage{booktabs}

\usepackage[a-1b]{pdfx}

\usepackage{hyperref} 

\title{\LARGE \bf Closed-loop Position Control of a Pediatric Soft Robotic Wearable Device for Upper Extremity Assistance}

\author{Caio Mucchiani,$^{1}$ Zhichao Liu,$^{1}$ Ipsita Sahin,$^{2}$ Jared Dube,$^{2}$ Linh Vu,$^{2}$ Elena Kokkoni,$^{2}$ Konstantinos Karydis$^{1}$
\thanks{$^{1}$ Dept. of Electrical and Computer Eng. and $^{2}$ Dept. of Bioengineering; Univ. of California, Riverside, 900 University Avenue, Riverside, CA 92521, USA. Email: {\tt\footnotesize\{caiocesr, zliu157, isahi001, jdube004, lvu035, elena.kokkoni, karydis\}@ucr.edu}. 
We gratefully acknowledge the support of NSF \# CMMI-2133084 and ARL \# W911NF-18-1-0266. 
Any opinions, findings, and conclusions or recommendations expressed in this material are those of the authors and do not necessarily reflect the views of the funding agencies.
}}

\begin{document}
\maketitle

\begin{abstract}
This work focuses on closed-loop control based on proprioceptive feedback for a pneumatically-actuated soft wearable device aimed at future support of infant reaching tasks. The device comprises two soft pneumatic actuators (one textile-based and one silicone-casted) actively controlling two degrees-of-freedom per arm (shoulder adduction/abduction and elbow flexion/extension, respectively). Inertial measurement units (IMUs) attached to the wearable device provide real-time joint angle feedback. Device kinematics analysis is informed by anthropometric data from infants (arm lengths) reported in the literature. Range of motion and muscle co-activation patterns in infant reaching are considered to derive desired trajectories for the device's end-effector. Then, a proportional-derivative controller is developed to regulate the pressure inside the actuators and in turn move the arm along desired setpoints within the reachable workspace. Experimental results on tracking desired arm trajectories using an engineered mannequin are presented, demonstrating that the proposed controller can help guide the mannequin's wrist to the desired setpoints.
\end{abstract}

\section{Introduction}

Innovation in pediatric assistive devices appears to lag compared to advances in devices for adults~\cite{fda,sanger2021opportunities, christy2016technology, arnold2020exploring}. 
Key differentiating factors (e.g., working with smaller body proportions that vary with time and supporting greater activity levels and longer use of devices compared to adults) make the development of pediatric devices extra challenging; especially those intended for use by infants (0-2 yrs of age) \cite{fda,sanger2021opportunities, christy2016technology}.
Specifically for upper extremity (UE) assistance, only a few passive devices are currently available (e.g., the Pediatric Wilmington Robotic Exoskeleton \cite{babik2016feasibility} and wearable Playskin \cite{lobo2016playskin}). 
Active assistance and feedback-based control, however, may be critical innovative features allowing for greater efficiency and higher precision when performing UE movements. 
In our previous work~\cite{kokkoni2020development} we introduced the first soft-actuated wearable for UE movement assistance designed specifically for infants, using silicone-based pneumatic actuators in a 4-DOF (two per limb) UE exosuit. 


\begin{figure}[!t]
\vspace{5pt}
\centering 
\includegraphics[trim={0 0 0 0.2cm},clip,width=1\columnwidth]{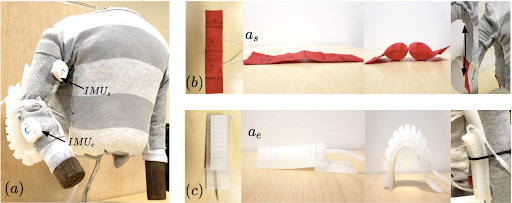}
\vspace{-21pt}
\caption{The current wearable prototype, featuring soft pneumatic actuators to generate motion about elbow and shoulder joints. (a) Testing with an engineered mannequin, using two IMUs one in the upper arm and one in the forearm. (b) Shoulder actuator design, with two channels and insert in the armpit. (c) Silicon-casted bending elbow actuator.}
\label{fig:actuators}
\vspace{-18pt}
\end{figure}

As active assistance becomes an important element for soft wearables, actuation needs to be considered carefully both in terms of the form/type of the actuators used and the type of control (open-loop, feedback) employed. Various types of soft actuators including 3D-printed~\cite{yap2016high,schaffner20183d}, silicone-casted~\cite{li2020high,kokkoni2020development} and textile~\cite{kim2021compact,nassour2020high,fu2022textiles,yap2017fully} ones have been developed and employed in the context of wearable devices~\cite{chen2021wearable} for adult rehabilitation~\cite{zahedi2021soft,natividad2020parallel}, strength augmentation~\cite{simpson2020upper,o2017soft} and fatigue reduction~\cite{nassour2020high,o2020inflatable}. 
%
%
From a closed-loop control design standpoint, a recent study proposed kinematic control of a pneumatic textile actuator to assist in shoulder motion~\cite{zhou2021kinematics}. Position and force control methods were explored in~\cite{abbasi2020position}. Data-driven approaches to characterizing bending angles and controlling soft actuators were also discussed~\cite{mohamed2020proposed,elgeneidy2018bending}. Position control methods for soft actuators often consider PID implementation \cite{ji2020design,morrow2016improving} and LQG \cite{hofer2018design}. 
In such feedback-based controllers (as in this work too), state estimation (e.g., end-effector position) of the device is required. 
To eventually make these devices useful in out-of-lab settings (where motion capture is not available) sensing methods for position estimation can utilize strain sensors \cite{yuen2018strain} and EMGs \cite{yap2017fully}, IMUs \cite{little2019imu}, flex sensors \cite{elgeneidy2018bending} and image analysis~\cite{li2021position,dechemi2021babynet}. Design for sensing in soft robotics is discussed in~\cite{tapia2020makesense}, whereas algorithmic considerations for infants' wearables are highlighted in~\cite{trujillo2017development}.
%

Key factors leading to successful pediatric assistive devices include weight, level of safety, and ease of operation~\cite{braito2018assessment,gonzalez2021robotic}. 
Herein we employ commercially-available IMUs for state feedback that are lightweight, enclosed, and directly transmit data that can be used by the user or within a software development pipeline (as herein). The current prototype (Fig.~\ref{fig:actuators}) comprises lightweight pneumatically-powered textile and silicone actuators for controlling motion about the shoulder and elbow joints for a single arm, respectively. A pneumatic control board to operate the actuators is mounted off-body at this stage of development.

The primary contribution of this paper is the development of a closed-loop pneumatic proportional-derivative (PD) controller based on proprioceptive feedback from two IMUs mounted on the upper arm and forearm links of an engineered mannequin. PD control with on/off switching of pneumatic valves \cite{hejrati2013accurate} has been found more robust in tracking higher frequencies than a PID controller. Other works have also used PD control for exoskeletons in rehabilitation \cite{meng2015design,chauhan2019series} and valve PWM switching for pneumatic position control \cite{taghizadeh2010multimodel}.

The controller is tasked to track arm trajectories deemed relevant to infant reaching according to the literature, and incorporates kinematics constraints related to how infants reach. The controller regulates the pneumatic pressure inside the two actuators to achieve end-effector (wrist) setpoint tracking. Experimental testing using the engineered mannequin shows that our proposed controller can work well to make the device's end-effector follow the desired trajectories.

\section{Problem Description}
\label{sec:problem}
\begin{figure}[!t]
\vspace{4pt}
\centering 
\includegraphics[width=0.75\columnwidth]{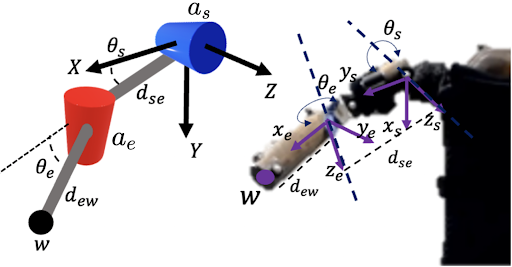}
\vspace{-10pt}
\caption{Reference frames, joint angles and links used in kinematics analysis.}
\label{fig:ik}
\vspace{-12pt}
\end{figure}

This work seeks to develop a \emph{closed-loop controller based on proprioceptive feedback for a pneumatically-actuated soft wearable device} that is analogous to a 2-DOF system, to support UE movement. We test the controller with a wooden mannequin (Fig.~\ref{fig:ik}) scaled based on infant anthropometric data. The torso of the mannequin is, for the purposes of this study, fixed in space. 
Consider the inertial coordinate system $P_i=(X,Y,Z)$, here taken to be aligned with the mannequin's right shoulder, and let the desired (calculated by inverse kinematics of a desired point $P_d$) shoulder and elbow joint angles (for the mannequin's arm to achieve) be $\theta_s$ and $\theta_e$, respectively.  
In contrast to rigid exoskeletons, these two joints are controlled via pneumatic actuators in our work. One pneumatic actuator ($a_s$) is responsible for shoulder adduction/abduction, while the other ($a_e$) controls elbow flexion/extension. 

The goal is to regulate the pneumatic actuators' internal pressure such that the final position of the wrist reaches $w=P_d$. To achieve so, we first generate desired trajectories for the joint angles in the form of quintic polynomials minimizing jerk in motion and respecting acceleration constraints. Every motion starts and finishes at rest, and motion constraints are shown in Table~\ref{table:constr}, where $\theta_s$, $\theta_e$ and $\ddot{\theta_s}$, $\ddot{\theta_e}$ are the shoulder and elbow joints position and acceleration, respectively. Then, we derive a Proportional-Derivative (PD) controller based on propioceptive feedback, namely two IMUs located at the upper arm and forearm of the mannequin.\footnote{~At this stage of development, the device operates at a quasistatic regime. As such, dynamic effects and disturbances have been excluded from modeling and control design and comprise part of future work.}

%
%
    

\begin{table}[htb]
\vspace{-6pt}
\centering
\caption{Kinematics Constraints}
\label{table:constr}
\vspace{-6pt}
\begin{tabular}{cccccc}
\toprule
\textbf{$\theta_{s_{MIN}}$} & \textbf{$\theta_{s_{MAX}}$} & \textbf{$\theta_{e_{MIN}}$}  & \textbf{$\theta_{e_{MAX}}$} & \textbf{$\ddot{\theta_s},\ddot{\theta_e}(t=0)$}  & \textbf{$\ddot{\theta_s},\ddot{\theta_e}(t=T)$}\\ \midrule
$0$ & $\pi/2$    & $0$ & $\pi/2$  & $0$ & $0$\\
\bottomrule
\end{tabular}
\vspace{-8pt}
\end{table}

\section{Technical Analysis}
\label{sec:methods}
This section contains the key technical contributions of the paper. We begin by performing a kinematics analysis of the wearable device. We then detail how to construct desired joint angle trajectories that are appropriate for infants. Lastly yet importantly, we develop a closed-loop PD controller for a pneumatically-actuated system based on IMU device egomotion feedback. 

\subsection{Device Kinematics Analysis}
Range of motion and acceptable workspace constraints for our wearable device were informed by relevant infant studies. 
First, the length of the upper arm ($d_{se}$) and forearm ($d_{ew}$) were chosen to be $d_{se}=d_{ew}=70$\;mm for a 7-month-old infant based on~\cite{sivan1983upper} (which considered 198 6-9 month-old infants). 
Second, our literature review on infant reaching kinematics~\cite{clifton1994multimodal,gonccalves2013development,bhat2006toy}, revealed an average reaching duration of about $1$\;sec, and average and peak velocities of $283$\;mm/s and $565.4$\;mm/s, respectively.
Third, reaching workspace information was obtained from a recent study with 10 infants affected by brachial plexus palsy, which assessed the UE reachable workspace of their affected and unaffected limbs via motion capture~\cite{richardson2022reachable}.
Lastly, information on infant movement control was derived by a study that assessed the effects of object size (among others) on reaching, and found that larger objects tend to be grasped in a feed-forward manner (symmetrical hand-speed profiles), in contrast to smaller objects where feedback control was present (asymmetrical hand-speed profiles)~\cite{berthier1996visual}. 
Figure~\ref{fig:work} visualizes these constraints in the form of the device's workspace (left panel) and desired configurations for the end-effector to reach (right panel).


With reference to Fig.~\ref{fig:ik}, the Denavit-Hartenberg parameters of the mechanism can be directly computed (Table~\ref{table:dh}). 

\begin{table}[htb]
\vspace{-6pt}
\centering
\caption{Denavit-Hartenberg Parameters}
\label{table:dh}
\vspace{-6pt}
\begin{tabular}{cccc}
\toprule
\textbf{$\theta_j$} & \textbf{$d_j$} & \textbf{$r_j$}  & \textbf{$\alpha_j$} \\ \midrule
$\theta_s$ & $0$    & $d_{se}$ & $\pi/2$                     \\
$\theta_e$ & $0$    & $d_{ew}$ & $0$ \\
\bottomrule
\end{tabular}
\end{table}
%
The homogeneous transformation matrix then is
 \begin{gather}
  \scalebox{1}{%
  $T =$
        $  \begin{bmatrix}
   c\theta_sc\theta_e &  -c\theta_ss\theta_e &  s\theta_s & c\theta_s(d_{se} + d_{ew}c\theta_e)\\
    s\theta_sc\theta_e &   -s\theta_ss\theta_e & -c\theta_s & s\theta_s(d_{se} + d_{ew}c\theta_e)\\
   s\theta_e & c\theta_e & 0 & d_{ew}s\theta_e \\
      0 & 0 & 0 &  1\\
  \end{bmatrix},
$
    \label{eq:ks}}
  \end{gather}

\noindent with orientation of the end-effector $\omega = [0,\theta_s,\theta_e]^T$. The forward kinematics leading to the end-effector position $[x,y,z]^T$ is given by the first three values of the last column of $T$, i.e.
 \begin{gather}
  \scalebox{1}{%
  $\begin{bmatrix}
   x \\
    y\\
    z\\
  \end{bmatrix}=$
   $\begin{bmatrix}
   c\theta_s(d_{se} + d_{ew}c\theta_e) \\
    s\theta_s(d_{se} + d_{ew}c\theta_e)\\
    d_{ew}s\theta_e\\
  \end{bmatrix}\enspace,
  $
   \label{eq:forw}}
 \end{gather}
and thus the joint angles are computed by
\begin{equation}
\theta_s = atan2(y,x), \quad
\theta_e = atan2(z,\sqrt{x^2+y^2}-d_{se})\enspace.
\end{equation}

The Jacobian matrix of the end-effector ($J_{ee}$) is 
\begin{gather}
  \scalebox{1}{%
  $J_{ee} =$
        $  \begin{bmatrix}
    -s\theta_s(d_{se} + d_{ew}c\theta_e) & -d_{ew}c\theta_ss\theta_e\\
      c\theta_s(d_{se} + d_{ew}c\theta_e) & -d_{ew}s\theta_ss\theta_e\\
    0 & d_{ew}c\theta_e \\
    0 & 0 \\
    0 & 1 \\
      1 &  0\\
  \end{bmatrix}\enspace.
$
    \label{eq:jacobian}}
  \end{gather}
%
$J_{ee}$ is non-square (Jacobian deficient) as the system is underactuated. In this case, since only two DOFs are controllable, we can separately analyze the first three rows of $J_{ee}$, disregarding the last two rows ($\omega_y$ and $\omega_z$) as they are redundant.

\begin{figure}[!t]
\vspace{6pt}
\centering 
\includegraphics[width=0.875\columnwidth]{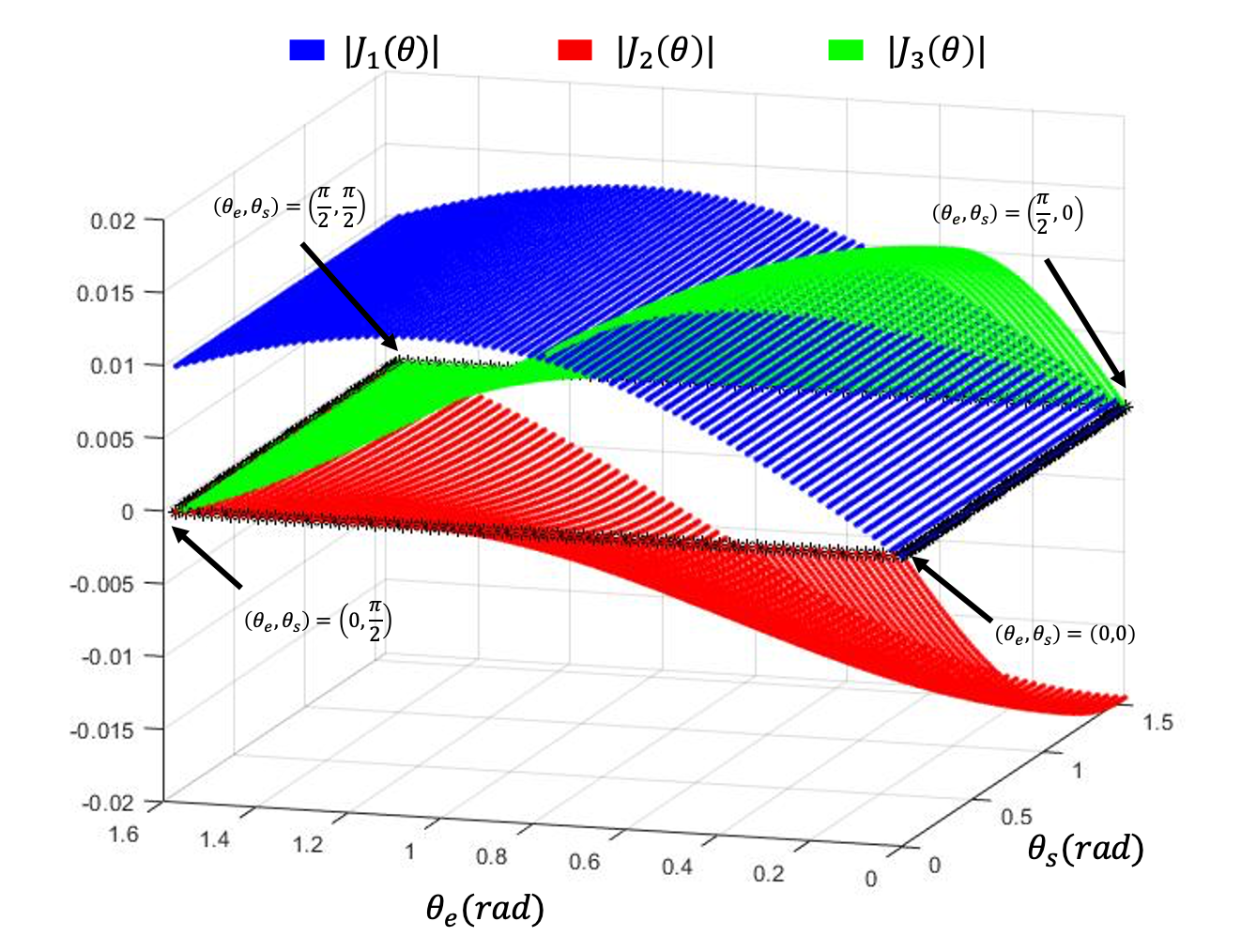}
\vspace{-9pt}
\caption{Jacobian determinant of the sub-matrices and singularity points, corresponding to shoulder and arm fully extended $(\theta_s,\theta_e)=\{(0,0),(\pi/2,0)\}$ and flexed elbow $(\theta_s,\theta_e)=\{(0,\pi/2),(\pi/2,\pi/2)\}$.}
\label{fig:singu}
\vspace{-18pt}
\end{figure}

Considering the square sub-matrices $J_1$, $J_2$ and $J_3$ as elimination of the third, second and first rows of $J_{ee}$ and taking the determinant and plotting against angles $\theta_s$ and $\theta_e$, we obtain regions of singularities (Fig.~\ref{fig:singu}). The workspace for the wearable can also be determined. Considering the position limits described in Table~\ref{table:constr}, a set of end-effector positions is calculated and approximated by alpha shapes~\cite{edelsbrunner1994three}. 

\subsection{Trajectory Generation}
Three distinct modes of actuation were considered based on previous work on muscle co-activation in infant reaching~\cite{spencer2000spatially} and prior testing of our initial prototype~\cite{kokkoni2020development}:

\begin{figure}[!t]
\vspace{6pt}
\centering 
\includegraphics[trim={0cm 0cm 0cm 0cm},clip,width=0.95\columnwidth]{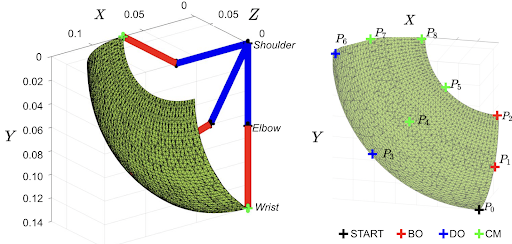}
\vspace{-6pt}
\caption{Wearable workspace (left) considering position constraints and average upper extremity and forearm measurements for infants and desired points (right) for trajectory generation.}
\label{fig:work}
\vspace{-18pt}
\end{figure}

\begin{itemize}
    \item Biceps only (BO): elbow actuator trajectory. 
   \item  Deltoid only (DO): shoulder actuator trajectory. 
    \item Combined Muscle (CM): elbow and shoulder actuators' trajectory combined.
\end{itemize}

\begin{table}[htb]
\vspace{-6pt}
\centering
\caption{Desired Points for Trajectories and Joint Angles}
\label{table:traj}
\vspace{-6pt}
\begin{tabular}{ccccc}
\toprule
\textbf{$P_d$} & $(X,Y,Z)$ & \textbf{$\theta_{d_s}$} & \textbf{$\theta_{d_e}$} & \textbf{$Type$} \\ \midrule
$P_0$ & $(0, 0, 0.1400)$ & $\pi/2$    & $0$  & $Start$ \\
$P_1$ & $(0, 0.1195, 0.0495)$ &$\pi/2$    & $\pi/4$               & $BO$    \\
$P_2$ & $(0, 0.0700, 0.0700)$  &$\pi/2$    & $\pi/2$                   & $BO$  \\
$P_3$ & $(0.0990, 0.0990, 0)$ &$\pi/4$    & $0$          &  $DO$       \\
$P_4$ & $(0.0845, 0.0845, 0.0495)$ &$\pi/4$    & $\pi/4$       & $CM$             \\
$P_5$ & $(0.0495, 0.0495, 0.0700)$ &$\pi/4$    & $\pi/2$        & $CM$             \\
$P_6$ & $(0.1400, 0, 0)$ &$0$    & $0$             & $DO$     \\
$P_7$ & $(0.1195, 0, 0.0495)$ &$0$    & $\pi/4$         & $CM$            \\
$P_8$ & $(0.0700, 0, 0.0700)$ &$0$    & $\pi/2$            & $CM$     \\
\bottomrule
\end{tabular}
\vspace{-9pt}
\end{table}

We consider eight distinct types of trajectories (two BO, two DO and four CM -- see Table~\ref{table:traj}). To create smooth and dynamically-feasible trajectories we consider quintic polynomial time scaling~\cite{lynch2017modern}, i.e.
\begin{equation}\label{eq:traj}
    P_d(t) = a_0t^5+a_1t^4+a_2t^3+a_3t^2+a_4t+a_5\enspace.
\end{equation}
Trajectories are constructed by taking the first and second derivative of~\eqref{eq:traj}, evaluating them and~\eqref{eq:traj} at the boundary conditions listed in Table~\ref{table:constr}, and solving a linear problem to identify coefficients $a_j,~ j=0,\ldots,5$. Desired configurations are visualized in the right panel of Fig.~\ref{fig:work}; corresponding numerical values are included in Table~\ref{table:traj}. The global reference frame at the shoulder joint is located at $P_0$ and $(X,Y,Z)$ represents the wrist coordinate.

\begin{figure}[!h]
\vspace{-6pt}
\centering 
\includegraphics[width=0.75\columnwidth]{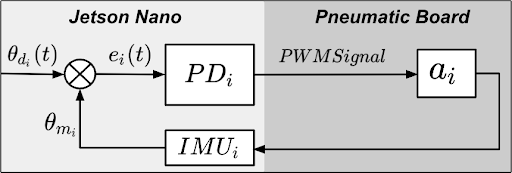}
\vspace{-9pt}
\caption{Schematic diagram of the PD controller for the wearable device.}
\label{fig:control}
\vspace{-9pt}
\end{figure}

\subsection{Pneumatic Feedback Control Design}
We develop two PD controllers ($PD_i$) to regulate the amount of pressure inside the actuators so that we can realize the aforementioned desired trajectories. The control structure is summarized in Fig.~\ref{fig:control}. Let the error signal be
 \begin{equation}
     e_i(t) = \theta_{d_i}(t) - \theta_{m_i}\enspace, \; i=s,e\enspace,
 \end{equation}
where $e_i(t)$ is the position error between the time-scaled desired joint angle values $\theta_{d_i}(t)$ (specified in Table~\ref{table:traj}) and angles measured online using the IMUs ($\theta_{m_i}$). The PD controller acts on the error producing a PWM $\%$ signal mapped to the actuators via a pneumatic control board (detailed in the next section). The controller attains the form  

\begin{equation}\label{eq:pd}
K_{gain_i}(t) = Kp_ie_i(t) + Kd_i\frac{d}{dt}e_i(t)\enspace, \;i=s,e \enspace,
\end{equation}
where $K_{gain_i}(t)$ are the gains translated to PWM signal for the actuators $a_i$ (with maximum value of $100$, meaning $100\%$ PWM), $Kp_i$ and $Kd_i$ are the proportional and derivative gains for each of the actuators $a_i$, respectively.

\begin{figure}[!t]
\vspace{6pt}
\centering 
\includegraphics[width=0.8\columnwidth]{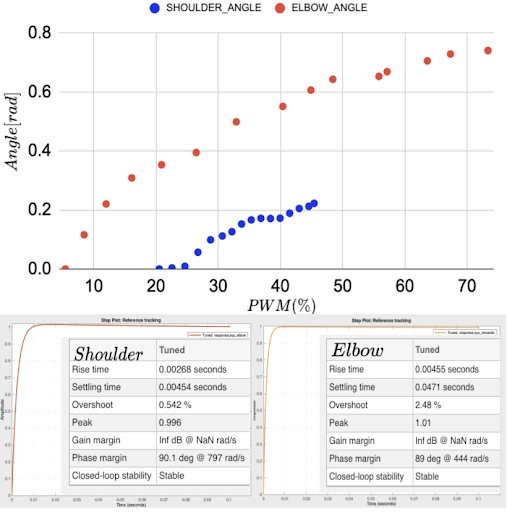}
\vspace{-6pt}
\caption{Averaged angle response based on different PWM values for both actuators (top) with sampling time $T_s=0.065 \;s$, and tuned PD controller based on the system model described in~\eqref{eq:tf}.}
\label{fig:tuning}
\vspace{-18pt}
\end{figure}

To approximate the system model, experimental data of PWM and angle variation (using a motion capture system) were collected using the same sample time ($T_s=0.0625\;s$) for both actuators and averaged (Fig.~\ref{fig:tuning}). Both systems were modeled as a spring-mass-damper, and Matlab (\textit{tfest}) was used to estimate the transfer functions as 
 \begin{equation}
     a_s = \frac{52.62}{s^2+15.57s+101.10} \\ ,\; \; \;
          a_e = \frac{16.11}{s^2+0.271s+36.88}
          \label{eq:tf}
 \end{equation}
with estimation fit of $90.58\%$ for the shoulder and $92.32\%$ for the elbow actuator, respectively.  The model in~\eqref{eq:tf} was tuned via Matlab (Fig.~\ref{fig:tuning}) using the \textit{pidTuner} Toolbox, resulting to $Kp_s=211$, $Kp_e=213$, $Kd_s=15$, and $Kd_e=27$.
 
\section{Experimental Testing and Results}
\label{sec:implement}

The wearable device and its pneumatic actuators are depicted in Fig.~\ref{fig:actuators}. The choice of components reflects the need for a device which minimizes discomfort (in terms of restriction of movement and weight) while also being functional and affordable. The experimental setup (hardware setup and software integration) for implementation as well as discussion of obtained results are described next.

\subsection{Experimental Setup}
A custom-made wooden
mannequin scaled according to infant anthropometric data~\cite{kokkoni2020development} was used to mount the wearable device 
(Fig.~\ref{fig:actuators}, left). For actuation, two different pneumatic designs were adopted (Fig.~\ref{fig:actuators}, right). The elbow actuator ($a_e$) is 
fabricated using 3D-printed casting molds and Smooth-On Dragon Skin 30 silicone~\cite{kokkoni2020development}. The shoulder actuator ($a_s$) was made via heat-sealable coated Oxford fabric \cite{sahin2022bidirect}, to allow for better mobility of the joint while still supporting the arm as compared to a silicone counterpart~\cite{kokkoni2020development}. 

\begin{figure}[!t]
\vspace{6pt}
\centering 
\includegraphics[width=0.975\columnwidth]{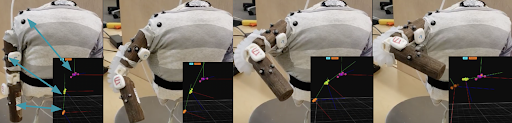}
\vspace{-8pt}
\caption{Experimental setup for the motion capture system and example of motion to point $P_8$ (from left to right).}
\label{fig:mocap}
\vspace{-18pt}
\end{figure}

The use of wireless IMUs \cite{little2019imu} intends to facilitate usability while also minimize restrictions on the movement on the device.
We have utilized the Mbientlab  
MetaMotionRl device which has 9-axis measurements of accelerometer, gyroscope and magnetometer, and provides sensor fusion streaming quaternions in real time using BLE communication. Two sensors (see Fig.~\ref{fig:actuators}) were placed at the midpoints of forearm ($IMU_s$) and upper arm ($IMU_e$) for estimating angles $\theta_s$ and $\theta_e$. 
For evaluation, a 12-camera Optitrack motion capture system (Fig.~\ref{fig:mocap}) was used for ground truth in experiments. 

\begin{figure}[htb]
\vspace{-6pt}
\centering 
\includegraphics[width=\columnwidth]{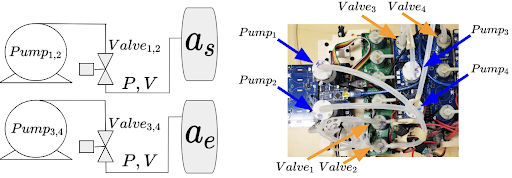}
\vspace{-18pt}
\caption{Pneumatic control board with pumps and valves distribution ($P,V$ indicates both pressure and vacuum can be sent to the actuators).}
\label{fig:pneu}
\vspace{-9pt}
\end{figure}

To implement the controller, an Nvidia Jetson Nano running ROS (Robot Operating System) communicated with the pneumatic control board (which uses the same layout as in~\cite{liu2021position}) through an Arduino Nano microcontroller running \textit{Rosserial}. The board (Fig.~\ref{fig:pneu}) has four pumps and four valves, with maximum pressure of $50\;$kPa and flow rate of $2 \;$l/min.


Control and IMU data acquisition modules are python scripts; communication modules to the pneumatic control board are in $C++$. Communication between both modules was done via ROS 
at a rate of $50$\;Hz. 


\subsection{Results and Discussion}\label{sec:results}
\begin{figure}[!t]
\vspace{6pt}
\centering 
\includegraphics[width=0.90\columnwidth]{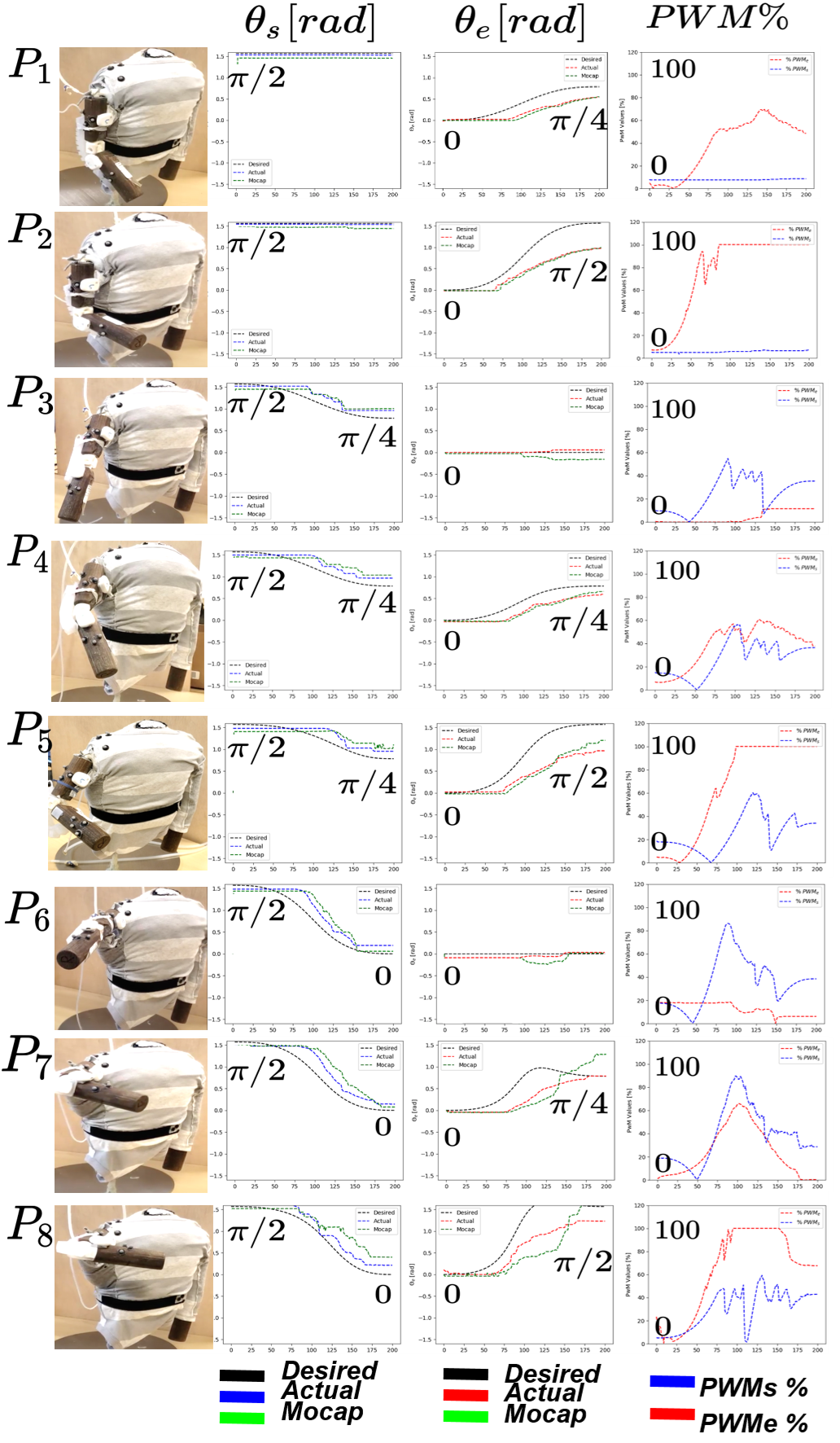}
\vspace{-10pt}
\caption{Experimental results for the spatial trajectories defined by Table \ref{table:traj}, there $P_0$ was the initial point in the trajectory. }
\label{fig:experiments}
\vspace{-9pt}
\end{figure}

A total of 64 experiments (eight per desired trajectory $P_i,~ i=1,\ldots,8$ as per Table~\ref{table:traj}) were performed to test our controller. Figure~\ref{fig:experiments} depicts obtained results. It can be readily verified that the controller performs well for most $BO$, $DO$ and $CM$ type trajectories, closely following the desired trajectories. 
Values for the $PWM \%$ were consistent with the expected behavior of the controller. A high (near $100 \%$) $PWM$ indicated more pneumatic effort to achieve the desired joint angle (usually required by the elbow actuator, since it has more mechanical resistance than the shoulder actuator, in $P_2$ and $P_5$ for instance), and no effort when no joint displacement was required (for shoulder in $BO$ and elbow in $DO$ type trajectories, in $P_2$ and $P_3$ for instance). 

\begin{figure}[!t]
\vspace{3pt}
\centering 
\includegraphics[width=\columnwidth]{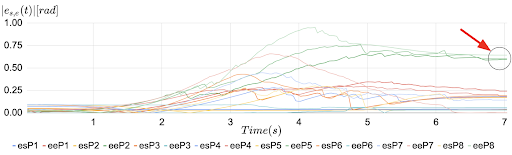}
\vspace{-18pt}
\caption{Average absolute position error of the shoulder and elbow actuators as a function of time for all trajectories. As it can be seen (red arrow), large steady state errors occurred for trajectories $P_2$, $P_5$ and $P_8$ as a full $\pi/2$ rotation of the elbow joint was required but not attained due to insufficient pressure from the pneumatic board. In addition, imprecise IMU readings may also have affected the controller input and thus the final position.}
\label{fig:error}
\vspace{-15pt}
\end{figure}

However, significant steady state error was observed for some trajectories requiring large elbow displacements $\theta_e >\pi/4$, such as for $P_2$, \;$P_5$, and $P_8$ (Fig.~\ref{fig:error}). 
Factors such as insufficient pressure from the pneumatic board (since this type of silicone-casted actuators requires higher pressure than the textile-based one) combined with the limited range of motion for the used actuators were the main contributors to steady state errors. Further, inability for the mannequin to maintain shoulder planar motion while performing elbow flexion/extension (interfering with the angular feedback from the IMUs), relative motion of the elbow actuator on the forearm and no fixture in one of its ends (and therefore allowing only active pushing and not pulling) were some of the contributing factors for the observed disparity. Although undesirable, these will inform future iterations of the elbow actuator to better address the desired range of motion, and also improve aspects such as actuator weight and mobility.

\section{Conclusion}
\label{sec:conclusion}

This work presented IMU-feedback-based closed-loop PD position control of a hybrid soft wearable device for infants, consisting of two pneumatic actuators assisting the motion of two DOF, at the shoulder and elbow joints. The controlled actuators were manufactured using different methods and materials, to allow for a better usability while also minimizing potential motion constraints on reaching movements. 
Device kinematics analysis along with anthropometric data from infant reaching informed the development of a trajectory generation method for the wearable to track smooth, dynamically-feasible trajectories. A PD controller was developed to regulate the amount of pressure inside the actuators for the wearable's end-effector to track the desired trajectories. 
Experimental results validated the proposed PD controller for most types of trajectories, but factors such as insufficient pressure from the actuators in achieving the final desired position were observed. 

Future work will build on top of the trajectory generation method and closed-loop controller developed herein by integrating 
force sensing to develop an impedance control method.  Additional sensing methods (such as flex sensors) can be combined with current sensors to investigate alternative control methods. 
Addition of an actuator to the wrist, use of actuators on both arms, and extension of the shoulder joint to a full 3-DOF motion will also be investigated.

\bibliographystyle{ieeetr}
\bibliography{root}

\begin{thebibliography}{10}

\bibitem{fda}
K.~Chowdhury, C.~Park, C.~Epps, and E.~Chen, ``{FDA'a Pediatric Device
  Consortia Grants Program: Narrowing the Gap in Pediatric Medical Device
  Development}.'' https://www.fda.gov/media/118152/download.
\newblock Accessed 2022-03-23.

\bibitem{sanger2021opportunities}
T.~Sanger, A.~Chang, W.~Feaster, S.~Taraman, N.~Afari, D.~Beauregard,
  B.~Dethlefs, T.~Ghere, M.~Kabeer, G.~Tolomiczenko, {\em et~al.},
  ``Opportunities for regulatory changes to promote pediatric device innovation
  in the united states: Joint recommendations from pediatric innovator
  roundtables,'' {\em IEEE Journal of Translational Engineering in Health and
  Medicine}, vol.~9, pp.~1--5, 2021.

\bibitem{christy2016technology}
J.~B. Christy, M.~A. Lobo, K.~Bjornson, S.~C. Dusing, E.~Field-Fote,
  M.~Gannotti, J.~C. Heathcock, M.~E. O'Neil, and J.~H. Rimmer, ``Technology
  for children with brain injury and motor disability: executive summary from
  research summit iv,'' {\em Pediatric Physical Therapy}, vol.~28, no.~4,
  pp.~483--489, 2016.

\bibitem{arnold2020exploring}
A.~J. Arnold, J.~L. Haworth, V.~O. Moran, A.~Abulhasan, N.~Steinbuch, and
  E.~Kokkoni, ``Exploring the unmet need for technology to promote motor
  ability in children younger than 5 years of age: a systematic review,'' {\em
  Archives of Rehabilitation Research and Clinical Translation}, vol.~2, no.~2,
  p.~100051, 2020.

\bibitem{babik2016feasibility}
I.~Babik, E.~Kokkoni, A.~B. Cunha, J.~C. Galloway, T.~Rahman, and M.~A. Lobo,
  ``Feasibility and effectiveness of a novel exoskeleton for an infant with arm
  movement impairments,'' {\em Pediatric Physical Therapy}, vol.~28, no.~3,
  p.~338, 2016.

\bibitem{lobo2016playskin}
M.~A. Lobo, J.~Koshy, M.~L. Hall, O.~Erol, H.~Cao, J.~M. Buckley, J.~C.
  Galloway, and J.~Higginson, ``Playskin lift: development and initial testing
  of an exoskeletal garment to assist upper extremity mobility and function,''
  {\em Physical therapy}, vol.~96, no.~3, pp.~390--399, 2016.

\bibitem{kokkoni2020development}
E.~Kokkoni, Z.~Liu, and K.~Karydis, ``Development of a soft robotic wearable
  device to assist infant reaching,'' {\em Journal of Engineering and Science
  in Medical Diagnostics and Therapy}, vol.~3, no.~2, 2020.

\bibitem{yap2016high}
H.~K. Yap, H.~Y. Ng, and C.-H. Yeow, ``High-force soft printable pneumatics for
  soft robotic applications,'' {\em Soft Robotics}, vol.~3, no.~3,
  pp.~144--158, 2016.

\bibitem{schaffner20183d}
M.~Schaffner, J.~A. Faber, L.~Pianegonda, P.~A. R{\"u}hs, F.~Coulter, and A.~R.
  Studart, ``3d printing of robotic soft actuators with programmable
  bioinspired architectures,'' {\em Nature communications}, vol.~9, no.~1,
  pp.~1--9, 2018.

\bibitem{li2020high}
H.~Li, J.~Yao, P.~Zhou, X.~Chen, Y.~Xu, and Y.~Zhao, ``High-force soft
  pneumatic actuators based on novel casting method for robotic applications,''
  {\em Sensors and Actuators A: Physical}, vol.~306, p.~111957, 2020.

\bibitem{kim2021compact}
W.~Kim, H.~Park, and J.~Kim, ``Compact flat fabric pneumatic artificial muscle
  (ffpam) for soft wearable robotic devices,'' {\em IEEE Robotics and
  Automation Letters}, vol.~6, no.~2, pp.~2603--2610, 2021.

\bibitem{nassour2020high}
J.~Nassour, F.~H. Hamker, and G.~Cheng, ``High-performance
  perpendicularly-enfolded-textile actuators for soft wearable robots: design
  and realization,'' {\em IEEE Transactions on Medical Robotics and Bionics},
  vol.~2, no.~3, pp.~309--319, 2020.

\bibitem{fu2022textiles}
C.~Fu, Z.~Xia, C.~Hurren, A.~Nilghaz, and X.~Wang, ``Textiles in soft robots:
  Current progress and future trends,'' {\em Biosensors and Bioelectronics},
  vol.~196, p.~113690, 2022.

\bibitem{yap2017fully}
H.~K. Yap, P.~M. Khin, T.~H. Koh, Y.~Sun, X.~Liang, J.~H. Lim, and C.-H. Yeow,
  ``A fully fabric-based bidirectional soft robotic glove for assistance and
  rehabilitation of hand impaired patients,'' {\em IEEE Robotics and Automation
  Letters}, vol.~2, no.~3, pp.~1383--1390, 2017.

\bibitem{chen2021wearable}
Y.~Chen, Y.~Yang, M.~Li, E.~Chen, W.~Mu, R.~Fisher, and R.~Yin, ``Wearable
  actuators: An overview,'' {\em Textiles}, vol.~1, no.~2, pp.~283--321, 2021.

\bibitem{zahedi2021soft}
A.~Zahedi, B.~Zhang, A.~Yi, and D.~Zhang, ``A soft exoskeleton for tremor
  suppression equipped with flexible semiactive actuator,'' {\em Soft
  robotics}, vol.~8, no.~4, pp.~432--447, 2021.

\bibitem{natividad2020parallel}
R.~F. Natividad, T.~Miller-Jackson, and R.~Y. Chen-Hua, ``A 2-dof shoulder
  exosuit driven by modular, pneumatic, fabric actuators,'' {\em IEEE
  Transactions on Medical Robotics and Bionics}, vol.~3, no.~1, pp.~166--178,
  2020.

\bibitem{simpson2020upper}
C.~Simpson, B.~Huerta, S.~Sketch, M.~Lansberg, E.~Hawkes, and A.~Okamura,
  ``Upper extremity exomuscle for shoulder abduction support,'' {\em IEEE
  Transactions on Medical Robotics and Bionics}, vol.~2, no.~3, pp.~474--484,
  2020.

\bibitem{o2017soft}
C.~T. O'Neill, N.~S. Phipps, L.~Cappello, S.~Paganoni, and C.~J. Walsh, ``A
  soft wearable robot for the shoulder: Design, characterization, and
  preliminary testing,'' in {\em IEEE International Conference on
  Rehabilitation Robotics (ICORR)}, pp.~1672--1678, 2017.

\bibitem{o2020inflatable}
C.~O’Neill, T.~Proietti, K.~Nuckols, M.~E. Clarke, C.~J. Hohimer,
  A.~Cloutier, D.~J. Lin, and C.~J. Walsh, ``Inflatable soft wearable robot for
  reducing therapist fatigue during upper extremity rehabilitation in severe
  stroke,'' {\em IEEE Robotics and Automation Letters}, vol.~5, no.~3,
  pp.~3899--3906, 2020.

\bibitem{zhou2021kinematics}
Y.~M. Zhou, C.~Hohimer, T.~Proietti, C.~T. O’Neill, and C.~J. Walsh,
  ``Kinematics-based control of an inflatable soft wearable robot for assisting
  the shoulder of industrial workers,'' {\em IEEE Robotics and Automation
  Letters}, vol.~6, no.~2, pp.~2155--2162, 2021.

\bibitem{abbasi2020position}
P.~Abbasi, M.~A. Nekoui, M.~Zareinejad, P.~Abbasi, and Z.~Azhang, ``Position
  and force control of a soft pneumatic actuator,'' {\em Soft Robotics},
  vol.~7, no.~5, pp.~550--563, 2020.

\bibitem{mohamed2020proposed}
M.~H. Mohamed, S.~H. Wagdy, M.~A. Atalla, A.~Rehan~Youssef, and S.~A. Maged,
  ``A proposed soft pneumatic actuator control based on angle estimation from
  data-driven model,'' {\em Journal of Engineering in Medicine}, vol.~234,
  no.~6, pp.~612--625, 2020.

\bibitem{elgeneidy2018bending}
K.~Elgeneidy, N.~Lohse, and M.~Jackson, ``Bending angle prediction and control
  of soft pneumatic actuators with embedded flex sensors--a data-driven
  approach,'' {\em Mechatronics}, vol.~50, pp.~234--247, 2018.

\bibitem{ji2020design}
Q.~Ji, X.~Zhang, M.~Chen, X.~V. Wang, L.~Wang, and L.~Feng, ``Design and closed
  loop control of a 3d printed soft actuator,'' in {\em IEEE International
  Conference on Automation Science and Engineering (CASE)}, pp.~842--848, 2020.

\bibitem{morrow2016improving}
J.~Morrow, H.-S. Shin, C.~Phillips-Grafflin, S.-H. Jang, J.~Torrey, R.~Larkins,
  S.~Dang, Y.-L. Park, and D.~Berenson, ``Improving soft pneumatic actuator
  fingers through integration of soft sensors, position and force control, and
  rigid fingernails,'' in {\em IEEE International Conference on Robotics and
  Automation (ICRA)}, pp.~5024--5031, 2016.

\bibitem{hofer2018design}
M.~Hofer and R.~D'Andrea, ``Design, modeling and control of a soft robotic
  arm,'' in {\em IEEE/RSJ International Conference on Intelligent Robots and
  Systems (IROS)}, pp.~1456--1463, 2018.

\bibitem{yuen2018strain}
M.~C. Yuen, R.~Kramer-Bottiglio, and J.~Paik, ``Strain sensor-embedded soft
  pneumatic actuators for extension and bending feedback,'' in {\em IEEE
  International Conference on Soft Robotics (RoboSoft)}, pp.~202--207, 2018.

\bibitem{little2019imu}
K.~Little, C.~W. Antuvan, M.~Xiloyannis, B.~A. De~Noronha, Y.~G. Kim, L.~Masia,
  and D.~Accoto, ``Imu-based assistance modulation in upper limb soft wearable
  exosuits,'' in {\em IEEE International Conference on Rehabilitation Robotics
  (ICORR)}, pp.~1197--1202, 2019.

\bibitem{li2021position}
D.~Li, V.~Dornadula, K.~Lin, and M.~Wehner, ``Position control for soft
  actuators, next steps toward inherently safe interaction,'' {\em
  Electronics}, vol.~10, no.~9, p.~1116, 2021.

\bibitem{dechemi2021babynet}
A.~Dechemi, V.~Bhakri, I.~Sahin, A.~Modi, J.~Mestas, P.~Peiris, D.~E.
  Barrundia, E.~Kokkoni, and K.~Karydis, ``Babynet: A lightweight network for
  infant reaching action recognition in unconstrained environments to support
  future pediatric rehabilitation applications,'' in {\em IEEE International
  Conference on Robot \& Human Interactive Communication (RO-MAN)},
  pp.~461--467, 2021.

\bibitem{tapia2020makesense}
J.~Tapia, E.~Knoop, M.~Mutn{\`y}, M.~A. Otaduy, and M.~B{\"a}cher, ``Makesense:
  Automated sensor design for proprioceptive soft robots,'' {\em Soft
  robotics}, vol.~7, no.~3, pp.~332--345, 2020.

\bibitem{trujillo2017development}
I.~A. Trujillo-Priego, C.~J. Lane, D.~L. Vanderbilt, W.~Deng, G.~E. Loeb,
  J.~Shida, and B.~A. Smith, ``Development of a wearable sensor algorithm to
  detect the quantity and kinematic characteristics of infant arm movement
  bouts produced across a full day in the natural environment,'' {\em
  Technologies}, vol.~5, no.~3, p.~39, 2017.

\bibitem{braito2018assessment}
I.~Braito, M.~Maselli, G.~Sgandurra, E.~Inguaggiato, E.~Beani, F.~Cecchi,
  G.~Cioni, and R.~Boyd, ``Assessment of upper limb use in children with
  typical development and neurodevelopmental disorders by inertial sensors: a
  systematic review,'' {\em Journal of Neuroengineering and Rehabilitation},
  vol.~15, no.~1, pp.~1--18, 2018.

\bibitem{gonzalez2021robotic}
A.~Gonzalez, L.~Garcia, J.~Kilby, and P.~McNair, ``Robotic devices for
  paediatric rehabilitation: a review of design features,'' {\em BioMedical
  Engineering OnLine}, vol.~20, no.~1, pp.~1--33, 2021.

\bibitem{hejrati2013accurate}
B.~Hejrati and F.~Najafi, ``Accurate pressure control of a pneumatic actuator
  with a novel pulse width modulation--sliding mode controller using a fast
  switching on/off valve,'' {\em Journal of Systems and Control Engineering},
  vol.~227, no.~2, pp.~230--242, 2013.

\bibitem{meng2015design}
W.~Meng, B.~Sheng, M.~Klinger, Q.~Liu, Z.~Zhou, and S.~Q. Xie, ``Design and
  control of a robotic wrist orthosis for joint rehabilitation,'' in {\em IEEE
  International Conference on Advanced Intelligent Mechatronics (AIM)},
  pp.~1235--1240, 2015.

\bibitem{chauhan2019series}
R.~J. Chauhan and P.~Ben-Tzvi, ``A series elastic actuator design and control
  in a linkage based hand exoskeleton,'' in {\em ASME Dynamic Systems and
  Control Conference}, vol.~59162, p.~V003T17A003, 2019.

\bibitem{taghizadeh2010multimodel}
M.~Taghizadeh, F.~Najafi, and A.~Ghaffari, ``Multimodel pd-control of a
  pneumatic actuator under variable loads,'' {\em The International Journal of
  Advanced Manufacturing Technology}, vol.~48, no.~5, pp.~655--662, 2010.

\bibitem{sivan1983upper}
Y.~Sivan, P.~Merlob, and S.~H. Reisner, ``Upper limb standards in newborns,''
  {\em American Journal of Diseases of Children}, vol.~137, no.~9,
  pp.~829--832, 1983.

\bibitem{clifton1994multimodal}
R.~K. Clifton, P.~Rochat, D.~J. Robin, and N.~E. Bertheir, ``Multimodal
  perception in the control of infant reaching.,'' {\em Journal of Experimental
  Psychology: Human perception and performance}, vol.~20, no.~4, p.~876, 1994.

\bibitem{gonccalves2013development}
R.~V. Gon{\c{c}}alves, E.~M. Figueiredo, C.~B. Mour{\~a}o, E.~A. Colosimo,
  S.~T. Fonseca, and M.~C. Mancini, ``Development of infant reaching behaviors:
  Kinematic changes in touching and hitting,'' {\em Infant Behavior and
  Development}, vol.~36, no.~4, pp.~825--832, 2013.

\bibitem{bhat2006toy}
A.~Bhat and J.~Galloway, ``Toy-oriented changes during early arm movements:
  Hand kinematics,'' {\em Infant Behavior and Development}, vol.~29, no.~3,
  pp.~358--372, 2006.

\bibitem{richardson2022reachable}
R.~T. Richardson, S.~A. Russo, R.~S. Chafetz, S.~Warshauer, E.~Nice, S.~H.
  Kozin, D.~A. Zlotolow, and J.~G. Richards, ``Reachable workspace with
  real-time motion capture feedback to quantify upper extremity function: A
  study on children with brachial plexus birth injury,'' {\em Journal of
  Biomechanics}, vol.~132, p.~110939, 2022.

\bibitem{berthier1996visual}
N.~E. Berthier, R.~K. Clifton, V.~Gullapalli, D.~D. McCall, and D.~J. Robin,
  ``Visual information and object size in the control of reaching,'' {\em
  Journal of Motor Behavior}, vol.~28, no.~3, pp.~187--197, 1996.

\bibitem{edelsbrunner1994three}
H.~Edelsbrunner and E.~P. M{\"u}cke, ``Three-dimensional alpha shapes,'' {\em
  ACM Transactions on Graphics}, vol.~13, no.~1, pp.~43--72, 1994.

\bibitem{spencer2000spatially}
J.~P. Spencer and E.~Thelen, ``Spatially specific changes in infants' muscle
  coactivity as they learn to reach,'' {\em Infancy}, vol.~1, no.~3,
  pp.~275--302, 2000.

\bibitem{lynch2017modern}
K.~Lynch and F.~Park, {\em Modern Robotics}.
\newblock Cambridge University Press, 2017.

\bibitem{sahin2022bidirect}
I.~Sahin, J.~Dube, C.~Mucchiani, K.~Karydis, and E.~Kokkoni, ``A bidirectional
  fabric-based pneumatic actuator for the infant shoulder: Design and
  comparative kinematic analysis,'' in {\em IEEE International Conference on
  Robot \& Human Interactive Communication (RO-MAN)}, 2022.

\bibitem{liu2021position}
Z.~Liu and K.~Karydis, ``Position control and variable-height trajectory
  tracking of a soft pneumatic legged robot,'' in {\em IEEE/RSJ International
  Conference on Intelligent Robots and Systems (IROS)}, pp.~1708--1709, 2021.

\end{thebibliography}

\end{document}